\documentclass{article}

\PassOptionsToPackage{numbers}{natbib}

\usepackage[preprint]{neurips_2022}

\usepackage[utf8]{inputenc} % allow utf-8 input
\usepackage[T1]{fontenc}    % use 8-bit T1 fonts
\usepackage{hyperref}       % hyperlinks
\usepackage{url}            % simple URL typesetting
\usepackage{booktabs}       % professional-quality tables
\usepackage{amsfonts}       % blackboard math symbols
\usepackage{nicefrac}       % compact symbols for 1/2, etc.
\usepackage{microtype}      % microtypography
\usepackage{xcolor}         % colors
\usepackage{graphicx}
\usepackage{enumitem}
\usepackage{multirow}
\usepackage{multicol}
\usepackage{amsmath}
\usepackage{wrapfig}
\usepackage{listings}
\usepackage{caption}
\usepackage{subcaption}
\usepackage{tipa}

\title{Regeneration Learning: A Learning Paradigm for Data Generation}

\author{
  Xu Tan$^1$\thanks{Corresponding author: Xu Tan, \texttt{xuta@microsoft.com}}~, ~Tao Qin$^1$, ~Jiang Bian$^1$, ~Tie-Yan Liu$^1$, ~Yoshua Bengio$^2$\\
$^1$Microsoft Research ~~$^2$Mila \& University of Montreal\\
$^1$\texttt{\{xuta,taoqin,jiabia,tyliu\}@microsoft.com}\\ $^2$\texttt{yoshua.bengio@mila.quebec}\\
}

\begin{document}

\maketitle

\begin{abstract}
Machine learning methods for conditional data generation usually build a mapping from source conditional data $X$ to target data $Y$. The target $Y$ (e.g., text, speech, music, image, video) is usually high-dimensional and complex, and contains information that does not exist in source data, which hinders effective and efficient learning on the source-target mapping. In this paper, we present a learning paradigm called \textit{regeneration learning} for data generation, which first generates $Y'$ (an abstraction/representation of $Y$) from $X$ and then generates $Y$ from $Y'$. During training, $Y'$ is obtained from $Y$ through either handcrafted rules or self-supervised learning and is used to learn $X \rightarrow Y'$ and $Y' \rightarrow Y$. Regeneration learning extends the concept of representation learning to data generation tasks, and can be regarded as a \textit{counterpart} of traditional representation learning, since 1) regeneration learning handles the abstraction ($Y'$) of the target data $Y$ for data generation while traditional representation learning handles the abstraction ($X'$) of source data $X$ for data understanding; 2) both the processes of $Y'\rightarrow Y$ in regeneration learning and $X\rightarrow X'$ in representation learning can be learned in a self-supervised way (e.g., pre-training); 3) both the mappings from $X$ to $Y'$ in regeneration learning and from $X'$ to $Y$ in representation learning are simpler than the direct mapping from $X$ to $Y$. We show that regeneration learning can be a widely-used paradigm for data generation (e.g., text generation, speech recognition, speech synthesis, music composition, image generation, and video generation) and can provide valuable insights into developing data generation methods.
\end{abstract}

\section{Introduction}
\subsection{Data Understanding and Generation}
Typical machine learning tasks, in the field of natural language processing~\cite{manning1999foundations,collobert2011natural,vaswani2017attention,devlin2018bert,brown2020language}, speech~\cite{benesty2008springer,hinton2012deep,oord2016wavenet,wang2017tacotron,tan2022naturalspeech}, computer vision~\cite{forsyth2011computer,szeliski2010computer,krizhevsky2012imagenet,he2016deep,goodfellow2014generative}, and etc, usually handle a mapping from source data $X$ to target data $Y$. For example, $X$ is image and $Y$ is class label in image classification~\cite{deng2009imagenet}; $X$ is style tag and $Y$ is sentence in style-controlled text generation~\cite{mckeown1992text}; $X$ is text and $Y$ is speech in text-to-speech synthesis~\cite{tan2021survey,tan2022naturalspeech}. 

\begin{figure} [t!]
\small
\begin{minipage}[b]{1.0\linewidth} 
	\centering
	\begin{subfigure}[b]{0.24\textwidth}
    \includegraphics[page=1,width=1.0\columnwidth,trim=0.0cm 0.8cm 8.5cm 0.8cm,clip=true]{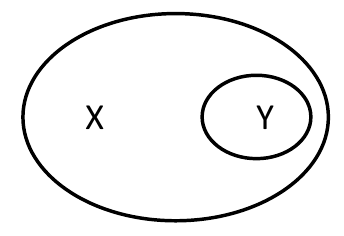} 
    \caption{Data understanding}
    \label{tab_task_information_understand}
    \end{subfigure}
    \begin{subfigure}[b]{0.35\textwidth}
    \includegraphics[width=1.0\columnwidth,trim=3.5cm 0.8cm 3.5cm 0.8cm,clip=true]{fig/data_information.pdf} 
    \caption{Data understanding/generation}
    \label{tab_task_information_both}
    \end{subfigure}
    \begin{subfigure}[b]{0.24\textwidth}
    \includegraphics[width=1.0\columnwidth,trim=8.5cm 0.8cm 0cm 0.8cm,clip=true]{fig/data_information.pdf}
    \caption{Data generation}
    \label{tab_task_information_generate}
    \end{subfigure}
\end{minipage} 
\begin{minipage}[b]{1.0\linewidth} 
\small
\vspace{0.2cm}
	\centering
	\begin{tabular}{ l  l l }
		\toprule
	    Types  & Information  & Tasks  \\
		\midrule
		\multirow{2}{*}{Understanding} & \multirow{2}{*}{$X \gg Y$} & image classification, objective detection  \\
		& & sentence classification, reading comprehension \\
		\midrule
		\multirow{2}{*}{Generation} & \multirow{2}{*}{$X \ll Y$} &  \multirow{2}{*}{text generation or image synthesis from ID/class} \\
		& & \\
		\midrule
		\multirow{2}{*}{Understanding/Generation} & \multirow{2}{*}{$X\not\gg Y$ and $X\not\ll Y$}  &  text to speech, automatic speech recognition, \\& & text to image generation, talking-head synthesis	 \\
		\bottomrule
	\end{tabular}
\end{minipage}
\caption{Three types of tasks in machine learning and the relative information between source $X$ and target $Y$.}
\label{tab_task_information_table}
\end{figure}

Depending on the relative amount of information that $X$ and $Y$ contain, these mappings can be divided into data understanding~\cite{krizhevsky2012imagenet,devlin2018bert}, data generation~\cite{goodfellow2014generative,brown2020language}, and the combination of data understanding and generation~\cite{bahdanau2014neural,hassan2018achieving,graves2012sequence,chan2016listen,tan2022naturalspeech}. Figure~\ref{tab_task_information_table} shows the three types of tasks and the relative information between $X$ and $Y$:
\begin{itemize}[leftmargin=*]
    \item Data understanding tasks, in which $X$ contains much more information than $Y$ (e.g., image classification~\cite{deng2009imagenet,krizhevsky2012imagenet}, objective detection~\cite{girshick2015fast,redmon2016you}, sentence classification~\cite{zhang2015sensitivity}, machine reading comprehension~\cite{rajpurkar2016squad}). 
    \item Data generation tasks, in which $Y$ contains much more information than $X$ (e.g., text generation~\cite{brown2020language} or image synthesis~\cite{goodfellow2014generative,kingma2013auto} from class label).
    \item Data understanding/generation tasks, in which $X$ contains no significantly more or less information than $Y$ (e.g., image transfer~\cite{zhu2017unpaired}, text-to-image synthesis~\cite{ramesh2021zero,ramesh2022hierarchical,yu2022scaling,saharia2022photorealistic,chang2023muse}, neural machine translation~\cite{bahdanau2014neural,hassan2018achieving}, text-to-speech synthesis~\cite{tan2022naturalspeech,tan2021survey}, automatic speech recognition~\cite{hinton2012deep}). In this case, we need both data understanding capability on the source $X$ and data generation capability on the target $Y$. 
\end{itemize}

The information mismatch between $X$ and $Y$ leads to different strategies for solving different tasks. For data understanding tasks, $X$ is usually high-dimensional, complex, and redundant compared to $Y$, and the key is to learn highly abstractive or discriminative representations (sometimes need to remove unnecessary information) for $X$ in order to better predict $Y$. Thus, representation learning~\cite{bengio2013representation}\footnote{One of the most impactful conferences in deep learning is ICLR, which is short for ``The International Conference on Learning Representations''.} and especially self-supervised pre-training~\cite{devlin2018bert,brown2020language} have become some of the hottest topics in deep learning research in the past years. For data generation tasks, $Y$ is usually high-dimensional, complex, and redundant compared to $X$, and the key is how to better represent the distribution of $Y$ and better generate $Y$ from $X$. For data understanding/generation tasks, they need the capability in both understanding and generation, i.e., extract good representations from $X$ and fully generate the information in $Y$.

\subsection{Challenges of Data Generation}
For the data generation tasks and the generation part of the data understanding/generation tasks (we call both the two types as data generation tasks in the remaining of this paper), they face distinctive challenges that cannot be addressed by the traditional formulation of representation learning. 
\begin{itemize}[leftmargin=*]
    \item First, for the generation tasks where $Y$ contains much more complex information than $X$, the generation models face severe one-to-many mapping problems (ill-posed or ill-condition problem)~\cite{bertero1988ill}, which increases the learning difficulty. For example, in class-conditioned image generation, a class label ``dog'' can correspond to different images that contain dogs. Incorrect modeling on the ill-posed problem could result in overfitting on the training set and poor generalization on the test set. 
    
    \item Second, for the generation tasks (e.g., speech recognition~\cite{hinton2012deep}, speech synthesis~\cite{tan2021survey}, talking-head video synthesis~\cite{thies2020neural}) where $X$ contains no significant more or less information than $Y$, there are two situations: 1) the mapping between $X$ and $Y$ is not one-to-one (e.g., multiple words with the same pronunciation can correspond to one speech segment in automatic speech recognition, and multiple speech with different speaking rates can correspond to one text sequence in text-to-speech synthesis), which faces the same problem mentioned above; 2) there are some spurious correlations between $X$ and $Y$, e.g., the speaking timber in source speech has no correlation with the head pose in target video~\cite{chen2019hierarchical} for talking-head video synthesis, and some information in target melody has no correlation with source lyric in lyric-to-melody generation~\cite{ju2021telemelody}. Fitting these spurious correlations can be harmful for generation in inference. 
\end{itemize}

\subsection{Why Regeneration Learning}
Some generative models (e.g., GANs~\cite{goodfellow2014generative}, VAEs~\cite{kingma2013auto}, autoregressive models~\cite{oord2016wavenet,brown2020language}, normalizing flows~\cite{rezende2015variational,kingma2018glow}, diffusion models~\cite{ho2020denoising}) have achieved rapid progress in a variety of data generation tasks. Ideally, as long as generative models are powerful enough, they can fit any complex data distribution. However, in practice, they cannot model the complex distribution and one-to-many mapping well due to many reasons, such as too complicated data mapping, too heavy computation cost, and data sparsity, etc. In analogy to data understanding tasks, although learning a powerful model (e.g., CNN or Transformer) to directly classify source data into target labels would ideally achieve very good accuracy, it still suffers from low accuracy, and some advanced representation learning methods such as large-scale self-supervised pre-training~\cite{devlin2018bert} can greatly boost the accuracy.  

In this paper, we present a learning paradigm called  \textit{regeneration learning} for data generation tasks. Instead of directly generating target data $Y$ from source data $X$, regeneration learning first generates $Y'$ (a representation of $Y$) from $X$ and then generates $Y$ from $Y'$. Regeneration learning extends the concept of representation learning to data generation tasks and learns a good representation ($Y'$) of the target data $Y$ to ease the generation: 1) $X\rightarrow Y'$ mapping will be less one-to-many than $X\rightarrow Y$ since $Y'$ is a compact/representative version of $Y$; 2) $Y'\rightarrow Y$ mapping can be learned in a self-supervised way ($Y'$ is obtained from $Y$) and can be empowered by large-scale pre-training that is similar to that in traditional representation learning for data understanding tasks (e.g., BERT~\cite{devlin2018bert}).

In the rest of this paper, we first introduce the basic formulation of regeneration learning and its connection to other learning methods and paradigms in Section~\ref{sec_reg_method}, then summarize the applications of regeneration learning in Section~\ref{sec_reg_app}, and finally list some research opportunities on regeneration learning in Section~\ref{sec_reg_opp}.

\section{Formulations of Regeneration Learning}
\label{sec_reg_method}

In this section, we introduce regeneration learning, which leverages intermediate representations of target data $Y$ to bridge the information mismatch between $X$ and $Y$. There are three steps in regeneration learning: 
\begin{itemize}[leftmargin=*]
\item Step 1: Convert target data $Y$ into an abstractive/representative version $Y'$. 
\item Step 2: Learn a model to generate $Y'$ from source data $X$.
\item Step 3: Learn another model to generate $Y$ from $Y'$. 
\end{itemize}
This learning paradigm is called \textit{regeneration learning} due to the following reasons: 1) Literally, it has two generation steps that first generate $Y'$ and then generate $Y$ (i.e., regenerate), which is in analogy to what ``represent'' is to ``present'' in representation learning\footnote{Strictly speaking, ``represent'' in representation learning  means ``using one thing to signify another thing'', which is different from ``re-present'' that means ``to present again''. However, the meaning of ``represent'' has a close relation to ``re-represent'': suppose you use $X'$ to signify $X$ (i.e., represent $X$ with $X'$ and thus $X'$ is the representation of $X$), and then in this case, you actually re-present $X$ using $X'$.}. 2) Metaphorically,  in analogy to what ``represent'' in representation learning means, i.e., ``using one thing to signify another thing'', the word ``regenerate'' in regeneration learning means to generate one thing to signify another thing (i.e., generate $Y'$ to signify $Y$). Regeneration learning has several advantages: 1) the mapping from $Y'$ to $Y$ can be learned in a self-supervised way, which is much more data-efficient; 2) the mapping from $X$ to $Y'$ is simpler than the direct mapping from $X$ to $Y$. 

We first introduce the three steps in regeneration learning in section~\ref{sec_reg_method_basic} and discuss the connections of regeneration learning to existing methods in Section~\ref{sec_reg_connect_method} and the relationships between regeneration learning and representation learning in Section~\ref{sec_reg_connect_rep}.

\subsection{Basic Formulation}
\label{sec_reg_method_basic}

%\paragraph{Principles of $Y \rightarrow Y'$ Conversion} 
\paragraph{Generate $Y'$ from Target $Y$.} There are three principles when converting $Y$ to $Y'$: 1) $Y'$ should be more abstractive and representative than $Y$; 2) the removed information from $Y$ to $Y'$ has no or little correlation with source data $X$, i.e., $Y'$ is a compact version of $Y$ but still maintains its correlation with $X$; 3) the conversion from $Y$ to $Y'$ should be easy, e.g., processed by simple transformation or extraction tools, or at least not relying on labeled data if model learning is needed. According to the above principles, there are different ways to convert $Y$ into $Y'$, as shown in the ``Basic Formulation'' in Table~\ref{tab_typical_data}: 

\begin{table}[t!]
\footnotesize
	\centering
	\begin{tabular}{l l l l l}
	\toprule
	Formulation & Category &  Method & Data Conversion ($Y\rightarrow Y'$) \\
	\midrule
	\multirow{10}{*}{Basic} &  \multirow{5}{*}{Explicit} & Fourier Transformation & Speech/Image (e.g, Wave$\rightarrow$ Spectrogram) \\
    &&  Grapheme-to-Phoneme & Text (e.g., learning$\rightarrow'$l\textipa{3:}rni\textipa{N}) \\
    &&  Music Analysis & Music (MIDI$\rightarrow$Chord/Rhythm) \\
    &&  3D Image Analysis & Image (Face to 3D Co-efficient) \\
    &&  Down Sampling & Speech/Image (e.g., 256*256$\rightarrow$64*64)  \\
	\cmidrule{2-4}
    &  \multirow{4}{*}{Implicit}  & Analysis-by-Synthesis & \multirow{4}{*}{Image/Speech/Text ($Y\rightarrow Z$)} \\
    && VAE  & \\
    && VQ-VAE/VQ-GAN & \\
    && DiffusionAE &  \\
 	\midrule
        \midrule
    \multirow{3}{*}{Extended} & Factorization & AR & Image/Speech/Text ($Y\rightarrow Y_{1:t}$) \\
     &  Diffusion & DDPM & Image/Speech/Text ($Y_{0}\rightarrow Y_{t}$)\\
     &  Latent Diffusion & VAE + DDPM & Image/Speech/Text ($Y\rightarrow Z_0$, $Z_{0}\rightarrow Z_{t}$)\\
	\bottomrule
	\end{tabular}
\vspace{0.3cm}
\caption{Different methods for $Y \rightarrow Y'$ conversion.}
\label{tab_typical_data}
\end{table}

\begin{itemize}[leftmargin=*]
\item Explicit transformation. We can convert $Y$ into $Y'$ with some explicit transformation methods: 1) Mathematical transformation such as Fourier or Wavelet transformation. For example, we can convert a speech waveform into a sequence of linear-scale or mel-scale spectrograms using short-time Fourier transformation (STFT)~\cite{wang2017tacotron,shen2018natural}. 2) Modality transformation such as grapheme-to-phoneme conversion. For example, we can convert a text/character/grapheme sequence into a phoneme sequence using the grapheme-to-phoneme conversion model~\cite{sun2019token}. 3) Data analysis. For example, we can extract music templates (chord, rhythm, etc) from a melody sequence using some music analysis tools to get the abstraction of the music~\cite{ju2021telemelody} or extract the 3D face parameters from a face image~\cite{ling2022stableface} using a 3D face model~\cite{blanz1999morphable}. 4) Downsampling. For example, we can simply down-sample an image from $256*256$ resolution to $64*64$ or a speech sequence from 48kHz sampling rate to 24kHz. These transformation methods are usually built on well-established methods or tools, and the $Y'$ transformed from $Y$ is usually in an explicit data format.  

\item Implicit transformation. Different from the explicit transformation that converts $Y$ into $Y'$ with explicit data format using some well-built rules, algorithms, or models, end-to-end learning achieves this conversion by learning an intermediate and implicit representation through analysis-by-synthesis pipeline or reconstruction. Some commonly-used models include auto-encoder (AE), denoising auto-encoder (DAE), variational auto-encoder (VAE)~\cite{kingma2013auto}, vector-quantized auto-encoder (VQ-VAE)~\cite{van2017neural,razavi2019generating}, diffusion autoencoder (DiffsuionAE)~\cite{preechakul2022diffusion}, etc. Beyond the learned encoder that converts $Y$ into $Y'$, they can additionally learn a decoder that converts $Y'$ back to $Y$, which can be used in the third step of regeneration learning.   
\end{itemize}

\paragraph{Generate $Y'$ from Source $X$.} The generation from $X$ to $Y'$ can be approached by any machine learning method, similar to those used to model $X\rightarrow Y$ (e.g., Autoregressive Model, GAN, VAE, Flow, Diffusion). Since $Y'$ is extracted from $Y$ satisfying the above three principles, the task of $X\rightarrow Y'$ is easier than that of $X\rightarrow Y$. For example, generating mel-spectrograms from text is much simpler than generating waveform in text-to-speech synthesis, generating 3D face parameters from the speech is much simpler than generating face image in talking-head video synthesis, and generating representation sequence (latent code sequence) from source condition is much simpler than generating pixel-level images in the conditional image or video generation.

\paragraph{Generate Target $Y$ from $Y'$.} Since we can easily extract $Y'$ from $Y$ by tools or models without relying on any paired training data (i.e., $X$ and $Y$), we can train a model to predict $Y$ from $Y'$ in a self-supervised way, where the paired training data $(Y', Y)$ can be collected in large scale without much cost. We can also conduct pre-training to enhance the capability of $Y'\rightarrow Y$ by only using a large amount of unpaired data $Y$. Note that when using auto-encoding methods to convert $Y$ into $Y'$, we can already get a decoder that converts $Y'$ to $Y$, without the need to train another model.

\paragraph{Infer $Y$ from $X$ via $Y'$.} We discuss how to combine the two generation steps $X\rightarrow Y'$ and $Y'\rightarrow Y$ together to generate $Y$ from $X$. Note that since $X\rightarrow Y$ mapping is one-to-many, the prediction of $Y$ from $X$ is distribution-wise, i.e., $Y\sim P(Y|X)$, instead of point-wise. Similarly, since $Y'$ is a compact version of $Y$ that incurs information loss, $Y'\rightarrow Y$ is also a one-to-many mapping and should also be modeled in a distribution-wise way, i.e., $Y\sim P(Y|Y')$. Furthermore, although the ill-posed mapping problem in $X\rightarrow Y'$ is largely alleviated compared to that in $X\rightarrow Y$, the $X\rightarrow Y'$ mapping is still one-to-many in general, and thus should be modeled in a distribution-wise way too, i.e., $Y'\sim P(Y'|X)$. On the other hand, $Y\rightarrow Y'$ is usually point-wise since it is converted by a deterministic function. But it can also be distribution-wise, such as using a VAE encoder to get the mean and variance of $Y'$ from $Y$. However, no matter whether $Y\rightarrow Y'$ is point-wise or distribution-wise, $X\rightarrow Y'$ and $Y'\rightarrow Y$ should be distribution-wise. We usually leverage deep generative models (e.g., autoregressive models, GANs, VAEs, normalizing flows, and denoising diffusion probabilistic models) to learn the conditional distributions $P(Y'|X)$ and $P(Y|Y')$. After that, given a data sample $X$, we can first sample $Y'$ from the conditional distribution $P(Y'|X)$, and then given the sampled data $Y'$, we can further sample $Y$ from the conditional distribution $P(Y|Y')$, i.e., $Y'\sim P(Y'|X)$, $Y\sim P(Y|Y')$.

\subsection{Connections to Other Methods}
\label{sec_reg_connect_method}

In this subsection, we discuss some methods that have connections to regeneration learning, including the methods that are basic and extended versions of regeneration learning, and that do not belong but are related to regeneration learning. 

\paragraph{Basic Versions of Regeneration Learning}
According to the formulation in Section~\ref{sec_reg_method_basic}, some methods can be regarded as the basic versions of regeneration learning, such as:
\begin{itemize}[leftmargin=*]
\item Template-based methods that extract a concrete template from target data and then regenerate target data from the template generated from source data~\cite{ju2021telemelody}.
\item Vocoding methods that convert speech waveform to spectrograms using Fourier transformation and regenerate waveform from the spectrograms that are generated from the source text or speech~\cite{wang2017tacotron,ren2019fastspeech,kong2020hifi}.
\item Grapheme/phoneme conversions that convert text/character sequence into phoneme sequence, and regenerate text/character sequence from the phoneme sequence that is generated from source speech~\cite{yuan2021decoupling}.
\item Auto-encoding methods that convert target data into representations and regenerate target data from representations that are generated from source data~\cite{van2017neural,razavi2019generating,ramesh2021zero,tan2022naturalspeech}.
\end{itemize}

\paragraph{Extended Versions of Regeneration Learning}
Some methods do not strictly follow the three-step definition of regeneration learning, but can be regarded as an extended version of regeneration learning. We show these $Y\rightarrow Y'$ conversion methods in ``Extended Formulation'' in Table~\ref{tab_typical_data} and introduce them as follows. 
\begin{itemize}[leftmargin=*]
\item In autoregressive generation, we use structure factorization to factorize the target data $Y$ into each step $Y_{1:t}$ in an autoregressive manner. We can regard a partial sequence $Y_{<t}$ as a simple/compact version of $Y_{<t+1}$, and all the intermediate partial sequences $Y_{<1}, Y_{<2}, ... Y_{<T}$ as the simple/compact versions of $Y_{1:T}$, where $T$ is the length of the target sequence $Y$. In this case, the regeneration contains multiple generation steps, and in each step, $Y_{<t+1}$ is generated from $Y_{<t}$. 

\item In denoising diffusion probabilistic models~\cite{ho2020denoising}, we can gradually add noise on original data $Y$ ($Y_0$) to get a diffused version of data $Y_t$~\cite{ho2020denoising}. Similar to autoregressive generation, the diffusion model is also an extended version of regeneration learning, where $Y_{t+1}$ can be regarded as a compact version of $Y_{t}$. Iterative-based non-autoregressive sequence generation~\cite{ghazvininejad2019mask} can also be regarded as regeneration learning following the same spirit.

\item There are some methods that combine auto-encoders with diffusion models~\cite{vahdat2021score,rombach2022high}, which first convert target data $Y$ into latent vector $Z$ (denoted as $Z_0$) using VAE or VQ-VAE, and then gradually add noise on $Z_0$ to get $Z_t$. They first generate $Z_0$ through the diffusion model and then regenerate target $Y$ using the decoder of VAE or VQ-VAE.  
\end{itemize}

\paragraph{Related to Regeneration Learning} We describe some methods that are related but do not belong to regeneration learning.
\begin{itemize}[leftmargin=*]
\item After $Y$ is generated, post-refine methods can be used to further improve the quality of generated $Y$. However, we do not regard them as regeneration learning, since they do not take the abstraction version of $Y$ but directly take $Y$ as the generation target. 
\item In the glancing mechanism~\cite{qian2020glancing,huang2021non}, some tokens from $Y$ are glanced at and taken as the input to predict $Y$. We also do not regard this mechanism as regeneration learning since there is no $Y'$ used in the learning process. 
\item In knowledge distillation~\cite{bucila2006model,hinton2015distilling,kim2016sequence}, the student takes the output of the teacher ($Y'$) as the learning target, which seems to be regeneration learning. However, knowledge distillation does not belong to regeneration learning due to several reasons. First, the student model in knowledge distillation does not regenerate $Y$ from $Y'$. Second, knowledge distillation can be distribution-wise~\cite{hinton2015distilling} or point-wise (e.g., sequence-level distillation~\cite{kim2016sequence}), and in distribution-wise distillation, $Y'$ is richer than $Y$, instead of an abstractive version of $Y$ as in regeneration learning.  
\end{itemize}
We will study the relation between regeneration learning and other iterative-based learning methods such as GFlowNet~\cite{bengio2021gflownet} for future work.  

\begin{figure} [t!]
\small
\begin{minipage}[b]{1.0\linewidth} 
\centering
	\begin{subfigure}[b]{0.1\textwidth}
    \includegraphics[page=2,width=1.0\columnwidth,trim=0.9cm 1.3cm 10cm 1.3cm,clip=true]{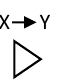} 
    \caption{}
    \label{tab_task_understand_info}
    \end{subfigure}
    \begin{subfigure}[b]{0.18\textwidth}
    \includegraphics[page=2,width=1.0\columnwidth,trim=2cm 1.3cm 8cm 1.3cm,clip=true]{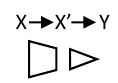} 
    \caption{}
    \label{tab_task_understand_info_both}
    \end{subfigure}
    \begin{subfigure}[b]{0.1\textwidth}
    \includegraphics[page=2,width=1.0\columnwidth,trim=4.4cm 1.3cm 6.5cm 1.3cm,clip=true]{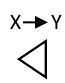} 
    \caption{}
    \label{tab_task_generate_info}
    \end{subfigure}
    \begin{subfigure}[b]{0.18\textwidth}
    \includegraphics[page=2,width=1.0\columnwidth,trim=5.9cm 1.3cm 4.1cm 1.3cm,clip=true]{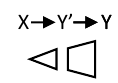} 
    \caption{}
    \label{tab_task_generate_info_both}
    \end{subfigure}
    \begin{subfigure}[b]{0.28\textwidth}
    \includegraphics[page=2,width=1.0\columnwidth,trim=8.2cm 1.3cm 0.7cm 1.3cm,clip=true]{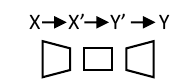} 
    \caption{}
    \label{tab_task_information_both}
    \end{subfigure}
\end{minipage}
Figure (a): Presentation ($X\rightarrow$). Figure (b): Representation ($X\rightarrow X'\rightarrow$). Figure (c): Generation ($\rightarrow Y$). Figure (d): Regeneration ($\rightarrow Y'\rightarrow Y$). Figure (e): Representation + Regeneration ($X\rightarrow X'\rightarrow Y'\rightarrow Y$).
\begin{minipage}[b]{1.0\linewidth} 
\footnotesize
	\centering
	\vspace{0.2cm}
	\begin{tabular}{l l l l l}
		\toprule
	    Paradigm & Original & Compact & Self-Supervised Learning & Easy Mapping \\
		\midrule
		(b) Representation Learning & $X$ & $X'$ & $X \rightarrow X'$ & $X' \rightarrow Y$ \\
		(d) Regeneration Learning & $Y$ & $Y'$ & $Y' \rightarrow Y$ & $X \rightarrow Y'$ \\
		\midrule
		(e) Combination & $X$, $Y$ & $X'$, $Y'$ & $X\rightarrow X'$, $Y'\rightarrow Y$ & $X'\rightarrow Y'$  \\
 		\bottomrule
	\end{tabular}
\end{minipage}
\caption{Comparison between regeneration learning and traditional representation learning on $X\rightarrow Y$. The triangle, trapezoid, and quadrangle in Figure (a)-(e) represent the information changes. Figure (a): Data understanding tasks, where $X$ contains more information than $Y$. Figure (b): Data understanding tasks with representation learning, where $X'$ is a more compact version of $X$. Figure (c): Data generation tasks, where $Y$ contains more information than $X$. Figure (d): Data generation tasks with regeneration learning, where $Y'$ is a more compact version of $Y$. Figure (e): Data understanding/generation tasks with the combination of representation and regeneration learning (e.g., sequence-to-sequence learning tasks such as speech to talking-head video synthesis), where $X'$ and $Y'$ contain a comparable amount of information and are compact versions of $X$ and $Y$ respectively (e.g., $X'$ could be speech representations learned by self-supervised models, and $Y'$ could be 3D coefficients extracted by 3D face model in talking-head video synthesis).}
\label{tab_representation_learning}
\end{figure}

\subsection{Regeneration Learning vs. Representation Learning}
\label{sec_reg_connect_rep}
Regeneration learning extends the concept of representation learning to data generation, and thus it can be regarded as a special type of representation learning for data generation. Furthermore, we can regard regeneration learning as a \textit{counterpart} of traditional representation learning, since 1) regeneration learning handles the abstraction ($Y'$) of the target data $Y$ for data generation, while traditional representation learning handles the abstraction ($X'$) of source data $X$ for data understanding; 2) both the processes of $Y'\rightarrow Y$ in regeneration learning and $X\rightarrow X'$ in traditional representation learning can be learned in a self-supervised way (e.g., pre-training); 3) both the mappings from $X$ to $Y'$ in regeneration learning and from $X'$ to $Y$ in traditional representation learning are simpler than the direct mapping from $X$ to $Y$. Figure~\ref{tab_representation_learning} shows the comparison between regeneration learning and traditional representation learning.

\section{Applications of Regeneration Learning}
\label{sec_reg_app}
A variety of tasks in conditional data generation (e.g., text generation, speech recognition, speech synthesis, music composition, image generation, and video generation) can benefit from this regeneration learning paradigm. We list some typical generation tasks in Table~\ref{tab_typical_task}, and introduce each task that leverages regeneration learning as follows.

\begin{table}[t!]
	\scriptsize
	\centering
	\begin{tabular}{l l l l l l}
		\toprule
	    Task  & $X$ & $Y$ & $Y'$ & $Y\rightarrow Y'$ \& $Y'\rightarrow Y$  \\
		\midrule
		Speech Synthesis & Text & Waveform & Spectrogram / Code & STFT \& Vocoder / Codec\\
		Speech Recognition & Speech & Character & Phoneme & G2P \& P2G \\
		Text Generation & Text/Knowledge & Text & Template & Text2Template \& Template2Text \\
		Lyric/Video to Melody & Lyric/Video & Melody & Music Template & Music Analysis \& Generation \\  
		Talking-Head Synthesis & Speech & Video & 3D Face Parameters & 3D Face Analysis \& Rendering \\
		Image/Video/Sound Generation & Class/Text & Image/Video/Sound & Latent Code & Codec Extraction \& Generation \\ 
 		\bottomrule
	\end{tabular}
	\vspace{0.3cm}
	\caption{ Typical data (text, speech, music, sound, image, and video) generation tasks that leverage regeneration learning.}
	\label{tab_typical_task}
\end{table}

\begin{itemize}[leftmargin=*]
\item \textit{Text-to-Speech Synthesis}. In text-to-speech synthesis (e.g., Tacotron 2~\cite{shen2018natural}, FastSpeech~\cite{ren2019fastspeech}), the target waveform is converted into mel-spectrogram sequence using short-time Fourier transformation (STFT), and an acoustic model is used to predict mel-specotrgram sequence from the source text, and a vocoder is used to generate waveform from predicted mel-spectrogram sequence. There are also some works~\cite{hayashi2020discretalk,liu2022delightfultts} using VQ-VAE to convert speech waveforms into continuous vectors or discrete tokens to bridge the mapping between text and speech.

\item \textit{Automatic Speech Recognition}. Some works~\cite{wang2021cascade,yuan2021decoupling,zhang2021decoupling} in automatic speech recognition first transform target character sequence into phoneme sequence using grapheme-to-phoneme (G2P) conversion, and use an acoustic model to predict phoneme sequence from source speech sequence, and then use another model to generate character sequence from predicted phoneme sequence.
 
\item \textit{Text Generation}. Some works~\cite{wiseman2018learning,yang2020improving} in text generation first convert a target sentence into a template or syntax sequence using template extraction or syntax parsing, and then predict template or syntax sequence from the source text or from scratch, and generate target text from the predicted template or syntax sequence.

\item \textit{Melody Generation}. Some works~\cite{ju2021telemelody,wu2020popmnet,dai2021controllable,zou2021melons,di2021video} in melody generation first extract music template from target melody sequence, and then predict the template from source conditions (e.g., lyrics, background videos, dancing videos) and generate a melody from the predicted template, such as TeleMelody~\cite{ju2021telemelody}.

\item \textit{Talking-Head Video Synthesis}. In talking-head video synthesis, some pipelines~\cite{thies2020neural,ji2021audio,yi2020audio,lahiri2021lipsync3d,song2021tacr,ling2022stableface,tang2022memories} for high-quality face synthesis usually extracts the 3D face parameters from the target face images through 3D face models~\cite{blanz1999morphable,thies2016face2face,feng2021deca}, and generates the 3D face parameters from source speech or text, and then generates the face images from the generated 3D face parameters. 

\item \textit{Image/Video/Sound Generation}. In image/video/sound generation, since the images or audios are too complex that incur too much computation, many works~\cite{ramesh2021zero,ding2021cogview,yan2021videogpt,rakhimov2020latent,borsos2022audiolm,yang2022diffsound,kreuk2022audiogen} first covert images/waveforms into discrete visual/audio tokens using VQ-VAE/VQ-GAN~\cite{van2017neural,esser2021taming,zeghidour2021soundstream} or into continuous hidden representations using CLIP~\cite{radford2021learning,ramesh2022hierarchical}, and then generate these visual/audio tokens or hidden either conditioning on text (e.g., DALL-E 1/2~\cite{ramesh2021zero,ramesh2022hierarchical}, StableDiffusion~\cite{rombach2022high}, AudioGen~\cite{kreuk2022audiogen} ) or from prefix/scratch (e.g., AudioLM~\cite{borsos2022audiolm}, VideoGPT~\cite{yan2021videogpt}, LVT~\cite{rakhimov2020latent}), and finally regenerate images/videos/audios from the generated tokens using the decoder in previous VQ-VAE/VQ-GAN or separate decoders such as GANs, VAEs, normalizing flows, or diffusion models.

\end{itemize}

Basically speaking, a conditional data generation task can leverage regeneration learning as long as they fit into some situations: 
\begin{itemize}[leftmargin=*]
\item The target data is too high-dimensional and complex to generate, or incurs too much computation cost, such as waveform generation in text-to-speech synthesis~\cite{shen2018natural,ren2019fastspeech,tan2021survey}, and image/video/sound generation~\cite{ramesh2021zero,ding2021cogview,yan2021videogpt,rakhimov2020latent,rombach2022high,kreuk2022audiogen}. In this case, converting target data $Y$ into more compact $Y'$ will greatly reduce the computation cost and free the model to focus more on how to generate high-level abstractive or semantic information of target data, but not on the minor details. 
\item The source data $X$ and target data $Y$ have too much uncorrelated information (i.e., $X \cap Y \ll X \cup Y$), such as lyric/video and melody in conditional melody generation~\cite{ju2021telemelody,wu2020popmnet,dai2021controllable,zou2021melons,di2021video}, speech and face images in talking-head video synthesis~\cite{thies2020neural,ji2021audio,yi2020audio,lahiri2021lipsync3d,song2021tacr,ling2022stableface}. Directly learning the mapping between $X$ and $Y$ would lead to overfitting. Thus, converting $Y$ into more compact $Y'$ will make the mapping between $X$ and $Y'$ less ill-posed and ease the model learning. 
\item There lack of paired $X$ and $Y$, and thus regeneration learning can be leveraged to train $Y'\rightarrow Y$ with large-scale self-supervised learning based on only target data $Y$. 
\end{itemize}

\section{Research Opportunities on Regeneration Learning}
\label{sec_reg_opp}
We discuss some research opportunities to make regeneration learning more powerful to solve a variety of data generation tasks, and list some corresponding research questions (RQ) both in each subsection and in Table~\ref{tab_research_question}, mainly from three perspectives: 1) how to get $Y'$; 2) how to learn the mapping $X\rightarrow Y'$ and $Y'\rightarrow Y$; 3) how to reduce the training-inference mismatch in regeneration learning pipeline.

\subsection{How to Get $Y'$} 
How to find an appropriate $Y'$ is important for $X\rightarrow Y'$ and $Y'\rightarrow Y$ mapping. For example, in text-to-speech synthesis, mel-spectrograms and mel-frequency cepstral coefficients (MFCCs) are both possible $Y'$ for target waveform $Y$. However, mel-spectrograms are demonstrated to be much better than MFCCs, since MFCCs are too abstractive that lose a lot of fine-grained information, thus making it difficult to reconstruct $Y$ from $Y'$. In the following, we list several possible research points to get a better $Y'$.

\paragraph{Better Implicit Learning.} Beyond using some handcrafted rules or well-developed tools to find $Y'$, we can automatically learn $Y'$ from $Y$. The typical methods are based on an analysis-by-synthesis pipeline, e.g., variational auto-encoder (VAE)~\cite{kingma2013auto} and vector-quantized variational auto-encoder (VQ-VAE)~\cite{van2017neural}. For example, in text-to-speech synthesis~\cite{cong2021glow,tan2022naturalspeech,liu2022delightfultts}, NaturalSpeech~\cite{tan2022naturalspeech} leverages VAE and DelightfulTTS 2~\cite{liu2022delightfultts} leverages VQ-VAE to learn intermediate representations $Y'$ from speech waveform $Y$ and reconstruct $Y$ from $Y'$. The motivation is that the commonly-used intermediate representations (e.g., mel-spectrograms) are extracted by SFTF algorithms that may not be optimal, while those learned by VAE could be better representations of waveform and ease the generation process. In the image and video domain, VQ-VAE~\cite{van2017neural} is widely used to learn discrete visual tokens to represent images and videos. However, some other works try to further improve the discrete token extraction of VQ-VAE by introducing hierarchical learning (e.g., VQ-VAE-2~\cite{razavi2019generating}), adding adversarial loss (e.g., VQ-GAN~\cite{esser2021taming}) and perceptual loss (e.g.,  PECO~\cite{dong2021peco}),  residual quantizers~\cite{gray1984vector,zeghidour2021soundstream,defossez2022high}, or leverage diffusion models to learn the hidden representation~\cite{preechakul2022diffusion}. 
\begin{itemize}[leftmargin=*]
    \item \textbf{RQ1}: How to design better analysis-by-synthesis methods (beyond VAE, VQ-VAE, DiffusionAE, etc) to learn $Y'$? 
    \item \textbf{RQ2}: How to design better learning paradigms other than analysis-by-synthesis to learn $Y'$? 
\end{itemize}

\begin{table}[t!]
	\scriptsize
	\centering
	\begin{tabular}{l l l}
		\toprule
        Perspective & ID & Research Questions \\
        \midrule
        \multirow{6}{*}{$Y\rightarrow Y'$} & 1 & How to design better analysis-by-synthesis methods (beyond VAE, VQ-VAE, DiffusionAE, etc) to learn $Y'$? \\
        & 2 & How to design better learning paradigms other than analysis-by-synthesis to learn $Y'$? \\
        & 3 &  How to leverage unpaired data $Y$ and/or paired data $(X, Y)$ to learn $Y'$? \\
        & 4 & How to better trade off the difficulty between $X\rightarrow Y'$ and $Y'\rightarrow Y$ mappings when learning $Y'$? \\   
        & 5 & How to disentangle semantic meaning and perceptual details to learn a more semantic instead of detailed $Y'$?  \\
        & 6 & How to determine the discrete or continuous format of $Y'$ for each data generation task?  \\
        \midrule
        \multirow{3}{*}{\shortstack{$X\rightarrow Y'$\\ $Y'\rightarrow Y$}} & 7 & How to design better generative models to learn $X\rightarrow Y'$ and $Y'\rightarrow Y$ mapping? \\ 
        & 8 & How to leverage the assumption of semantic conversion and detail rendering to design better methods? \\
        & 9 & How to leverage large-scale self-supervised learning for $Y'\rightarrow Y$ mapping? \\
        \midrule
 	$X\rightarrow Y'\rightarrow Y$ & 10 & How to reduce the training-inference mismatch in regeneration learning? \\
  \bottomrule
	\end{tabular}
	\vspace{0.3cm}
	\caption{Research questions on regeneration learning.}
	\label{tab_research_question}
\end{table}

\paragraph{Learning $Y'$ by Considering $X$.} An intuition is that the abstractive representation $Y'$ should not only depends on $Y$, but also be correlated to $X$ in order to facilitate the prediction of $Y'$ from $X$. To this end, we formulate a learning strategy as follows. We take the target data $Y$ as the input of the target encoder to get $Y'= f_Y(Y)$ and use the target decoder to reconstruct $Y\sim P(Y|Y')$. We further take source data $X$ as the input of the source encoder to get $X'=f_X(X)$ and encourage $X'$ and $Y'$ to contain information with each other by using some losses such as contrastive loss or L1/L2 loss: $||X'-f_{Y'}(Y')||^2_2$ and $||Y'-f_{X'}(X')||^2_2$. Additionally, we use a loss $-\log P(Y|f_{X'}(X'))$ to encourage the target decoder to get used to the $Y'$ predicted from $X'$, which can also address the problems mentioned in Section~\ref{sec_reg_opp_mismatch}. The total loss is as follows: 
\begin{equation}
L = -\log P(Y|Y') +||X'-f_{Y'}(Y')||^2_2 +||Y'-f_{X'}(X')||^2_2 -\log P(Y|f_{X'}(X')),  
\label{eq_learning_y}
\end{equation}
where $Y'=f_Y(Y)$ and $X'=f_X(X)$. We can leverage both unpaired data $Y$ and paired data $(X, Y)$ to learn $Y'$ according to Equation~\ref{eq_learning_y}. When using unpaired data $Y$ for learning, we only use the first loss term. 
We leave the theoretical and empirical investigation of Equation~\ref{eq_learning_y} as future work, but at the same time also welcome researchers to study this topic if you are interested. 

Except for leveraging paired data $(X, Y)$ to learn a good $Y'$, we can also leverage some prior knowledge about $X$ when choosing how to convert $Y$ into $Y'$. For example, in automatic speech recognition, we know that the source $X$ is speech and thus we convert text sequence $Y$ into phoneme sequence $Y'$ to ease the conversion from $X$ to $Y'$. In text-to-image synthesis (e.g., DALL-E 2~\cite{ramesh2022hierarchical}), the image is converted into CLIP~\cite{radford2021learning} features that are close to text representations to ease text-to-image conversion. 

However, there is a trade-off on the difficulties between $X\rightarrow Y'$ and $Y'\rightarrow Y$, since considering more $X$ when learning $Y'$ will ease the learning of $X\rightarrow Y'$ while making the learning of $Y'\rightarrow Y$ harder. We should make a good trade-off to get better overall performance.

\begin{itemize}[leftmargin=*]
    \item \textbf{RQ3}: How to leverage unpaired data $Y$ and/or paired data $(X, Y)$ to learn $Y'$? 
    \item \textbf{RQ4}: How to better trade off the difficulty between $X\rightarrow Y'$ and $Y'\rightarrow Y$ mappings when learning $Y'$?   
\end{itemize}

\paragraph{Disentangling Semantic Meaning from Perceptual Details.}
Generally speaking, $X\rightarrow Y'$ cares more about semantic mapping/conversion from source to target, while $Y'\rightarrow Y$ cares more about rendering perceptual details to obtain the target $Y$. For example, in text-to-image synthesis, $X\rightarrow Y'$ converts source text into discrete visual tokens that describe the semantic meanings of the target image, and $Y'\rightarrow Y$ renders the image details from visual tokens. How to design a learning mechanism to better disentangle the semantic meaning and perceptual details and let $Y'$ focus on semantic meaning will be a good research opportunity. 
\begin{itemize}[leftmargin=*]
    \item \textbf{RQ5}: How to disentangle semantic meaning and perceptual details to learn a more semantic instead of detailed $Y'$?  
\end{itemize}

\paragraph{Discrete vs. Continuous $Y'$.}
Both discrete and continuous $Y'$ can be leveraged to bridge the mapping between $X$ and $Y$. For example, using similar VQ-VAE to quantize speech waveforms, DelightfulTTS 2~\cite{liu2022delightfultts} leverages continuous vectors as $Y'$ while DiscreTalk~\cite{hayashi2020discretalk} leverages discrete tokens as $Y'$. Which kinds of representations (discrete vs continuous) are better choices for $Y'$ is also an interesting point to investigate. 
\begin{itemize}[leftmargin=*]
    \item \textbf{RQ6}: How to determine the discrete or continuous format of $Y'$ for each data generation task?  
\end{itemize}

\subsection{How to Learn $X\rightarrow Y'$ and $Y'\rightarrow Y$}
Regeneration learning decomposes a conditional data generation task $X\rightarrow Y$ into data conversion and data rendering processes. The data conversion process converts source data $X$ into the target domain, which maintains the concrete semantics but does not necessarily contain fine-grained details. The data rendering process further renders the fine-grained details of the target data to achieve high-quality data generation. Roughly speaking, $X\rightarrow Y'$ undertakes more on the role of data conversion, while $Y'\rightarrow Y$ undertakes more on the role of data rendering.

How to design better training methods on $X\rightarrow Y'$ and $Y'\rightarrow Y$ would be important for the final performance of $X\rightarrow Y$ mapping in regeneration learning. Advanced generative models such as autoregressive models~\cite{oord2016wavenet,brown2020language}, VAEs~\cite{kingma2013auto}, GANs~\cite{goodfellow2014generative}, normalizing flows~\cite{dinh2014nice,rezende2015variational}, and diffusion models~\cite{sohl2015deep,ho2020denoising} can play an important role. Furthermore, considering $Y'\rightarrow Y$ mapping can usually be learned through large-scale self-supervised methods, we should leverage more unpaired data when learning $Y'\rightarrow Y$.

\begin{itemize}[leftmargin=*]
    \item \textbf{RQ7}: How to design better generative models to learn $X\rightarrow Y'$ and $Y'\rightarrow Y$ mapping? 
    \item \textbf{RQ8}: How to leverage the assumption of semantic conversion in $X\rightarrow Y'$ and detail rendering in $Y'\rightarrow Y$ to design better methods? 
    \item \textbf{RQ9}: How to leverage large-scale self-supervised learning for $Y'\rightarrow Y$ mapping? 
\end{itemize}

\subsection{How to Reduce Training-Inference Mismatch in $X\rightarrow Y'\rightarrow Y$}
\label{sec_reg_opp_mismatch}
The model of $Y'\rightarrow Y$ is trained in a self-supervised way, where $Y'$ is extracted from $Y$ in training. However, $Y'$ is predicted from $X$ in inference, which causes the training-inference mismatch. How to reduce the mismatch is important to ensure the performance of this cascaded system ($X\rightarrow Y'$ and $Y'\rightarrow Y$). A straightforward way is to design an end-to-end optimization method between $X\rightarrow Y'$ and $Y'\rightarrow Y$, but still maintain $Y'$ as an intermediate representation. For example, NaturalSpeech~\cite{tan2022naturalspeech} leverages VAEs and normalizing flows with bidirectional prior/posterior optimization to achieve end-to-end learning. Can we design other methods to reduce the training-inference mismatch in regeneration learning? 

\begin{itemize}[leftmargin=*]
    \item \textbf{RQ10}: How to reduce the training-inference mismatch in regeneration learning? 
\end{itemize}

\section{Conclusion}
In this paper, we present a learning paradigm called regeneration learning for data generation tasks. Literally, it means generating the data two times: first generates an intermediate representation $Y'$ from source data $X$, and then generates a target data $Y$ from $Y'$. Metaphorically, it means generating  $Y'$ as intermediate representations to signify $Y$, in analogy to representation learning. Regeneration learning can be regarded as a counterpart of traditional representation learning: regeneration learning handles the abstraction ($Y'$) of the target data $Y$ for data generation while representation learning handles the abstraction ($X'$) of source data $X$ for data understanding, and both the processes of $Y'\rightarrow Y$ in regeneration learning and $X\rightarrow X'$ in representation learning can be learned in a self-supervised way (it is also a counterpart in literally: presentation$\rightarrow$representation vs. generation$\rightarrow$regeneration). We discuss the connections of regeneration learning to other methods, demonstrate a variety of data generation tasks that can benefit from regeneration learning, and further point out some research opportunities on regeneration learning. Regeneration learning can be a widely-used paradigm for high-quality data generation and can provide valuable insights into developing data generation methods.

\section*{Acknowledgements}
We thank our colleagues in the Machine Learning Group of Microsoft Research Asia for the discussions. We thank Qi Meng and Chang Liu for providing helpful insights and suggestions on this paper. 

%\newpage
\bibliography{neurips_2022}
\bibliographystyle{plainnat}

\end{document}